\documentclass[conference]{IEEEtran}
\usepackage{times}
 
\usepackage[numbers]{natbib}
\usepackage{multicol}
\usepackage[bookmarks=true]{hyperref}
\usepackage{graphicx, amsmath, mathtools}

\usepackage{amsmath,amsfonts}
\usepackage{amsthm}
\usepackage{algorithm}
\usepackage{algorithm}  
\usepackage{algorithm}  
\usepackage{algorithm}  
\usepackage{algpseudocode}
\usepackage{array}
\usepackage[caption=false,font=normalsize,labelfont=sf,textfont=sf]{subfig}
\usepackage{textcomp}
\usepackage{stfloats}
\usepackage{url}
\usepackage{verbatim}
\usepackage{graphicx}
\usepackage{mathtools}
\usepackage{float}
\usepackage{xcolor}
\usepackage{makecell}
\usepackage{booktabs}  
\usepackage{array, caption, capt-of, cuted}     
\usepackage[font=small,labelfont=bf]{caption}
\usepackage{arydshln} 
\usepackage{fontawesome5}

\usepackage{xcolor}

\newcommand{\bx}{\mathbf{x}}
\newcommand{\cH}{\mathcal{H}}
\newcommand{\MMD}{\text{MMD}}

\newcommand{\Homeo}{\operatorname{Homeo}}
\newcommand{\Id}{\operatorname{Id}}
\newcommand{\Lip}{\operatorname{Lip}}

\newcommand{\R}{\mathbb{R}}

\newcommand{\espace}{E}

\newtheorem{definition}{Definition}
\newtheorem{prop}{Proposition}

\begin{document}

\title{Asymptotically Optimal Ergodic Coverage on Generalized Motion Fields\vspace{-15pt}}




\author{\authorblockN{Christian Hughes\authorrefmark{1},
Yilang Liu\authorrefmark{1},
Yanis Lahrach\authorrefmark{5},
Julia Engdahl\authorrefmark{2},
Houston Warren\authorrefmark{3},
Darrick Lee\authorrefmark{4}, 
Fabio Ramos\authorrefmark{3}, \\
Travis Miles\authorrefmark{2}, and
Ian Abraham\authorrefmark{1}\authorrefmark{3}}
\authorblockA{\authorrefmark{1}Yale University, \authorrefmark{5} Université Catholique de Louvain, \authorrefmark{2}Rutgers University, \authorrefmark{4}University of Edinburgh, \authorrefmark{3}University of Sydney}}




%

\makeatletter
        \let\@oldmaketitle\@maketitle%
        \renewcommand{\@maketitle}{
        \@oldmaketitle
      \centering  
      \includegraphics[width=\textwidth]{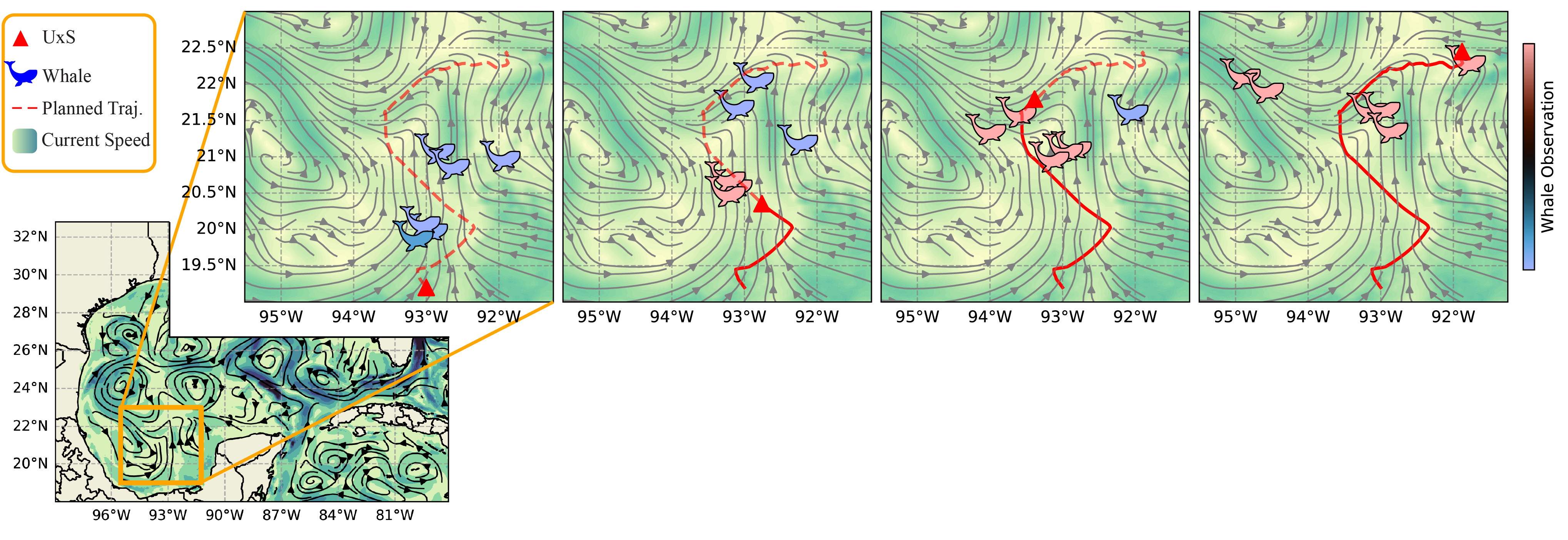}
      \vspace{-77pt}
      \captionsetup{margin={5.5cm,0cm}}
      \captionof{figure}{\textbf{Autonomous Migratory Tracking over the Gulf.} We propose a novel coverage trajectory planner that reasons about coverage over abstract domains that evolves according to flow fields. Here, the search domain corresponds to the whale population that flows with the ocean currents. The autonomous uncrewed underwater system (UxS) equipped with our proposed method integrates the flow model into guaranteeing coverage and observational frequency despite its under-actuation.}
      \label{fig:whales}
        \vspace{-5pt}
        \addtocounter{figure}{-1}%
        }
\makeatother

\maketitle

\let\thefootnote\relax\footnotetext{This work is supported by the Office of Naval Research grant N000142512440 and the National Science Foundation under award NSF FRR 2238066. Any opinions, findings, and conclusions or recommendations expressed in this material are those of the authors and do not necessarily reflect the views of the National Science Foundation or Office of Naval Research.
}

\addtocounter{figure}{1}%

\begin{abstract}
Autonomous robotic exploration in remote and extreme environments allows scientists to model complex transport phenomena and collective behaviors described by continuously deforming flow fields.
Although these environments are naturally modeled as time-varying domains, most adaptive exploration methods assume static environments and fail to provide adequate coverage or satisfy any formal guarantees. 
This is especially the case in oceanography where autonomous underwater systems (UxS) have highly restrictive compute and payload requirements that necessitate path planning methods that yield robust data collection strategies in open-loop and underactuated settings. 

In this work, to address the aforementioned issues, we propose to formulate adaptive search as an ergodic coverage problem and investigate certifying coverage in the ergodic sense over evolving domains with flow-induced dynamics. 
We expand upon recent work demonstrating maximum mean discrepancy (MMD) as a functional ergodic metric, and derive a flow-adaptive formulation that explicitly accounts for domain evolution within the coverage objective. 
We show that this approach preserves ergodic coverage guarantees in ambient flows and enables effective exploration in under-actuated, and even open-loop planning settings by integrating environment dynamics. 
Experiments validate that our method generalizes to diverse spatiotemporal processes including ocean exploration, and tracking human and cattle movement. Physical experiments on aerial and legged robotic platforms validate our ability to obtain ergodic coverage in non-convex, flow-restricted environments while respecting robot dynamics.  
\end{abstract}

\IEEEpeerreviewmaketitle

\begin{figure*}
    \centering
    \includegraphics[width=\linewidth]{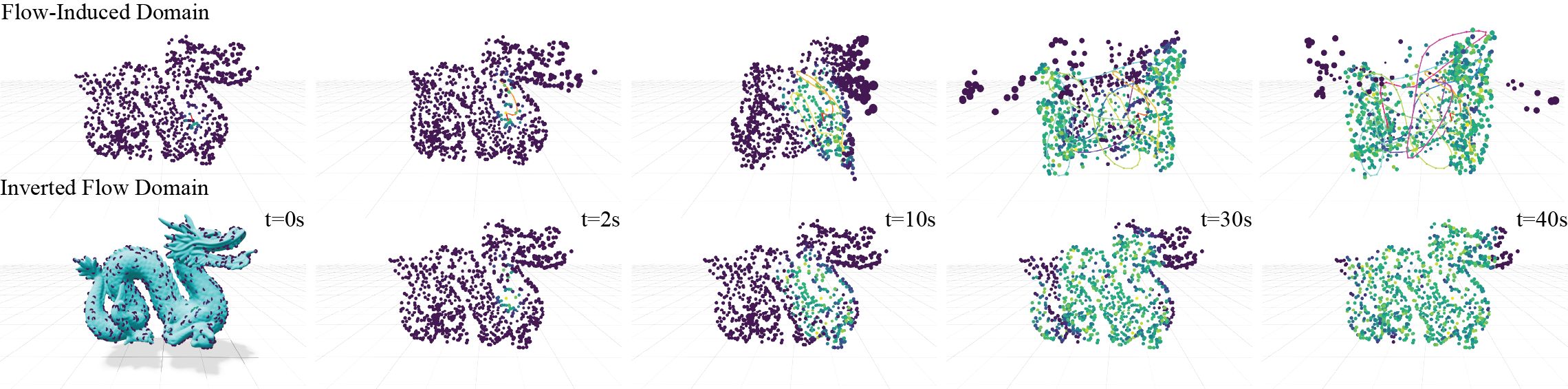}
    \caption{\textbf{Illustration of Ergodic Trajectory Optimization Over a Flow-Induced Domain.} The search domain is defined as a set of points uniformly drawn from the Stanford Dragon mesh. (Top) Shown is the evolution of the domain subject to a flow of a Duffing oscillator. Lighter green color denotes visitation by the trajectory rendered over time through the HSV colormap. (Bottom) Illustration of coverage fidelity on initial ambient domain. Our approach leverages a novel definition of ergodicity on the inverted flow map that facilitates computationally efficient open-loop trajectory optimization and establishes formal guarantees of coverage.  
    }
    \vspace{-12pt}
    \label{fig:bunny_search_flow}
\end{figure*}

\section{Introduction}

Autonomous exploration remains a pivotal challenge in robotics, particularly in remote and extreme environments where planning for data collection is a long-term and arduous process. 
Classical approaches, such as random sampling \cite{kavraki1996} and greedy information-gain strategies \cite{ivic2023}, excel at quickly identifying high-value regions but fall short in incorporating long-term planning and providing coverage guarantees. 
Alternatively, traditional coverage-based planning methods, such as grid-based exploration \cite{sung2023, bahnemann2021}, have achieved moderate success at search tasks in structured and well-known environments, but lack consideration for environment changes that structurally alter the search domain. 
The core difficulty of exploring an evolving domain lies in the challenge of balancing efficient long-horizon visitation proportionality with short-term information gain. 
Furthermore, energy-constrained and underactuated systems, such as UxS, require stable and robust long-duration trajectories to avoid the high computational cost of iterative planning \cite{schofield2007, creed2004}. 
Therefore, in this work, we seek to address the question: \textit{can robotic systems perform effective coverage-based search and exploration over domains subjected to flows?}

Recent ergodic search methods have emerged as promising tools for exploration that balance short-term information gain with long-term coverage guarantees.
Such methods construct exploration trajectories that maximize time-averaged visitation statistics proportional to a predefined information distribution \cite{mathew2011}, and, over sufficiently long time-horizons, allow robots to explore a domain such that their visitation asymptotically approaches a given search space's utility distribution.

Unfortunately, traditional ergodic search methods assume a priori knowledge of the search domain and underlying information distribution \cite{mathew2011, sun2024}, and few consider time-varying definitions of information in the formulation ~\cite{mavrommati2018}, nor do they consider time-varying domains in the search problem (i.e. where the search space changes over time).
While recent advancements have identified Maximum Mean Discrepancy (MMD), a two-sample statistical metric ~\cite{gretton2012}, as a kernel-based ergodic metric for exploration in arbitrary spatial domains \cite{hughes2024, lahrach2025}, its application has been largely restricted to stationary environments.

In this paper, we extend the ergodic MMD formulation to time-varying domains by defining ergodicity over discrete spatial samples that evolve according to a known flow field.
Unlike spectral methods, MMD's sample-based approach facilitates navigation in complex, non-convex geometries by decoupling the ergodic metric from a fixed coordinate structure. 
We derive a flow-adaptive variation of MMD that is capable of establishing asymptotic coverage guarantees in the reachable space over long-duration planning horizons. 
Our findings demonstrate that the proposed ergodic MMD framework effectively couples a robot's control with the underlying flow physics of the domain. 
We demonstrate our approach on a number of robotic exploration problems that require traversal over evolving search landscapes, including under-actuated and, more critically, open-loop exploration problems. 
In summary, our contributions are: 
\begin{enumerate}
    \item A theoretical formulation of ergodicity that maintains coverage guarantees for information distributions transported by time-varying vector fields; 
    \item A novel MMD-based optimization method that enables ergodic exploration in non-convex and flow-constrained domains; and
    \item Experimental demonstrations of our approach's versatility in search problems subjected to fluidic currents, animal migration, and human crowd flows.
\end{enumerate}

The remainder of this paper is structured as follows: Section~\ref{sec:related_work} discusses prior and related work. Section~\ref{sec:preliminaries} covers preliminary material on ergodicity, maximum-mean discrepancy, and ergodic control. Section \ref{sec:main} then derives the proposed time-varying ergodic maximum-mean discrepancy method and provides theoretical analysis on the approach. Lastly, Sections \ref{sec:results} - \ref{sec:conclusion} present a series of experimental results, limitations, and future research directions.

\section{Related Work}
\label{sec:related_work}

\subsection{Information-Theoretic Exploration}

Information-driven exploration has emerged as a crucial strategy for autonomous systems to efficiently reduce environmental uncertainty by optimizing trajectories according to the predicted information gain \cite{bourgault2002, amigoni2010}.
Early foundations of information maximization used probabilistic models to minimize uncertainty across the search space \cite{thrun1997}, while later approaches use Gaussian Processes to model information gain as the inverse of predicted variance \cite{jadidi2014, marchant2014_2}.
These methods are highly effective for static mapping and reconstruction tasks where sensor data may be sparse or noisy.

Recent advancements have shown improvements upon traditional information-theoretic methods using sparse environment sampling \cite{hollinger2014, caley2020}. 
These methods have also been modified to enable flow-tracking, which allows for exploration in dynamic systems \cite{marchant2014}. 
However, while successful in short-term tasks or those in which high-value regions are the sole priority, these methods often lack the long-horizon coverage guarantees necessary for comprehensive exploration. 
In this paper, we propose a flow-adaptive ergodic planner that provides theoretical guarantees for comprehensive coverage of the reachable space in non-stationary domains over long-horizon planning problems. 

\subsection{Ergodic Trajectory Optimization}



Ergodic trajectory planning methods seek to minimize the discrepancy between a robot's time-averaged visitation and a target spatial utility distribution. 
Early work in ergodic trajectory optimization demonstrated theoretical guarantees of full coverage of continuous-space domains over infinite time-horizons by minimizing a spectral ergodic metric based on Fourier basis functions \cite{abraham2020, mathew2011, scott2009}. 
However, these spectral methods are largely restricted to known, bounded Euclidean domains, which limits their utility in environments with non-convex geometries \cite{hughes2024, mathew2011}. 

Recent work has introduced Maximum Mean Discrepancy (MMD) as a functional ergodic metric for arbitrarily-structured domains \cite{hughes2024}. 
While this formulation enables exploration of complex and non-convex search spaces, current developments of this approach assume that the exploration space is static. 
We extend this approach by proposing a modified MMD metric that accounts for domain evolutions subject to approximated or known flow fields to enable simultaneous planning for current and future utility landscapes. 

\subsection{Flow Tracking}

When interacting with environments whose topography or information measure are time-variant, it is crucial that autonomous systems possess the ability to track movements within their exploration manifold. 
Lagrangian Coherent Structures (LCS) provide a principled framework for analyzing time-dependent flows by identifying material surfaces that guide optimal motion in environments such as multi-agent swarms and underwater currents \cite{short2014, hayat2024, haller2015, haller2001, senatore2008, peng2009}. 

Common approaches to manifold tracking involve approximating LCS by either estimating fluid velocity or tracking discrete points within the flow \cite{li2022, haller2002}. 
Because LCS are extensions of both stable and unstable manifolds that generate time-dependent flows, LCS inherently denote regions of their represented flow where escape events are more likely to occur \cite{paulos2015}. 
These unstable manifolds have also been shown to be approximable through local sampling of relevant regions \cite{michini2013}.
However, full flow-approximation of environments under complex flows is often computationally expensive or requires high-density sampling. 
Alternatively, some strategies restrict exploration to easily-approximated regions within the flow manifold to reduce information requirements \cite{kularatne2017}, but these methods risk missing important regions within the flow. 
In this paper, we leverage knowledge of the flow field to optimize trajectories that are robust to minor approximation irregularities and exploit ambient dynamics to maximize global coverage. 

\section{Preliminaries}
\label{sec:preliminaries}

In this section, we review the fundamental principles of ergodic search techniques, and outline the maximum mean discrepancy metric that we use to establish the time-varying domain coverage problem.

\subsection{Ergodicity}

    We define ergodicity by first considering a robot's state in discrete time as $x_t\in \mathcal{X} \subseteq \mathbb{R}^n$.
    Additionally, let us define $\mathbf{x} = \{x_0, \ldots, x_{T-1} \}$ as the trajectory of the robot for time-horizon $T \in \mathbb{N}$ following some control input $u_t \in \mathcal{U} \subseteq \mathbb{R}^m$ for $t = 0, \ldots, T-1$ and dynamics $x_{t+1} = f(x_t, u_t)$ from some initial condition $x_0$, and $f : \mathcal{X} \times \mathcal{U} \to \mathcal{X}$.
    Next, let us define the closed and bounded operational search space $\Omega$ where $\omega_t \in \Omega$ is a point in the operational search space. The function $g : \mathcal{X} \to \Omega$ then projects states $x_t \to \omega_t$, that is, $g$ defines where the robot is in the operational search space. 
    
    \begin{definition} \textit{(Time-Averaged Trajectory Distribution)}
        Assuming deterministic dynamics, the time-averaged trajectory distribution, e.g., the visitation histogram of a trajectory $\bx$ in $\Omega$, is given as 
        \begin{equation}
            \rho_{\bx,T}(\omega) = \frac{1}{T} \sum_{t=0}^{T-1} \delta[\omega- g(x_t)],
            \label{eq:time_averaged_distribution}
        \end{equation}   
        where $\delta$ is a Dirac-delta function.
    \end{definition}
    
    \begin{definition} \textit{(Ergodicity)} \label{def:ergodicity}
    ~\cite[Theorem 6.14]{walters_introduction_2000} Let $\mu : \Omega \to \mathbb{R}^+$ be a probability measure. A trajectory $\bx = \{ x_t\}_{t=0}^{T-1}$ in $\Omega$ is said to be ergodic with respect to $\mu$ if its time-averaged statistics $\rho_{\bx,T}(\omega)$ at the limit $T\to \infty$ converge weakly to $\mu$. In other words, 
    \begin{equation}
        \lim_{T\to \infty} \int_\Omega \phi(\omega) d \rho_{\bx, T}(\omega) = \int_\Omega \phi(\omega) d\mu(\omega)
    \end{equation}
    for all continuous functions $\phi \in \mathcal{C}(\Omega)$, where by definition,
    \begin{equation}
        \int_\Omega \phi(\omega) d \rho_{\bx, T}(\omega) = \frac{1}{T}\sum_{t=0}^{T-1} \phi(g \circ x_t)
    \end{equation}
    and the equality becomes
    \begin{equation}
        \lim_{T\to \infty} \frac{1}{T}\sum_{t=0}^{T-1} \phi(g \circ x_t) = \int_\Omega \phi(\omega) d\mu(\omega).
    \end{equation}
\end{definition}

    Intuitively, this statement of ergodicity suggests that a trajectory visits regions in $\Omega$ with probability proportional to the expected value of $\mu$ distributed over $\Omega$. 
    Establishing the ergodic property is ideal for information gathering and exploration that requires a robot to visit each place in the search domain.
    In general, measuring ergodicity for a robotic system does not occur over infinite time horizons and we typically consider problems where $T\ll \infty$ which we denote as sub-ergodic.

\subsection{Ergodic Control}

The concept of ergodicity can be rigorously quantified by defining a metric that evaluates the difference between a trajectory's time-averaged statistics and a given target distribution. More specifically, the distance to ergodicity~\eqref{def:ergodicity} is calculated by the following metric

\begin{equation} \label{eq:generic_ergodic_metric}
    \mathcal{E}_\mu(\mathbf{x}) = \Big\Vert \mathcal{F}[\rho_{\bx,T}] - \mathcal{F}[\mu] \Big\Vert_\mathcal{F}^2
\end{equation}
where $\mathcal{F}$ is a functional transform applied to the distributions $\rho$ and $\mu$, and $\Vert \cdot \Vert_\mathcal{F}$ is the associated norm in the transformed space.

Prior work has employed the Fourier transform as $\mathcal{F}$ \cite{mathew2011, lee2024} and more recent approaches adopt kernel-based techniques to approximate the $L^2$ ergodic metric~\cite{sun2024}. These metrics maintain the core property that $\mathcal{E}_\mu \to 0$ as $T \to \infty$ if and only if the trajectory $\mathbf{x}$ is ergodic with respect to the distribution $\mu$.

Given the definition of a robot, i.e., its dynamic constraints, the goal of ergodic control is to generate trajectories that are ergodic in a defined closed and bounded operational search space $\Omega$. To achieve ergodicity, the robot trajectory and controls are solved through an optimization problem that minimizes the ergodic metric. A general optimization problem can be expressed as:
\begin{align}
\label{eq:gen_ergodicity}
    \min_{\substack{\mathbf{x}, \mathbf{u}}}\,\, & \mathcal{E}_\mu(\mathbf{x})  + \sum_{t=0}^{T-1} \ell(x_t, u_t)\\         \text{subject to } & h_1(x_t, u_t) = 0 \,\, \forall t \in [0, T-1] \nonumber \\ & h_2(x_t, u_t) \le 0 \,\, \forall t \in [0,T-1] \nonumber
\end{align}
where $\mathbf{u}$ represents the control input sequence, $\ell$ is a running cost function, and the constraints $h_1$ and $h_2$ encapsulate dynamics, kinematic relations, and boundary conditions.

\subsection{Maximum Mean Discrepancy}

The ergodic metric in~\eqref{eq:generic_ergodic_metric} is obtained as a special case of a generic metric between distributions, where one of the distributions is the time-averaged trajectory distribution. 
Maximum mean discrepancy (MMD) is a two-sample statistical metric that measures the similarity between samples of two probability distributions using kernel mean embeddings in a Reproducing Kernel Hilbert Space (RKHS). Beyond expressivity, the advantage of MMD is that one does not necessarily need access to either underlying density function to compute the metric, only samples. 

To define the MMD metric, let $k: \Omega \times \Omega \to \mathbb{R}^{\ge 0}$ be a positive-definite kernel, with $\cH$ as the associated RKHS. The kernel mean embedding for a probability distribution $p$ is given by:
\begin{align}
    p \mapsto \mu_p = \mathbb{E}_{\omega \sim p}[ k(\omega, \cdot)].
\end{align}
When this mapping is injective, the kernel is called \emph{characteristic}~\cite{sriperumbudurUniversality2011}, which allows for a unique representation of $p$ in $\cH$.
The MMD between distributions $p$ and $q$ is expressed as
\begin{align}\label{eq:mmd_definition}
    \MMD_k^2(p, q) = \|\mu_p - \mu_q\|_{\cH}^2
\end{align}
which can be expanded to
\begin{align} \label{eq:expanded_mmd}
        \MMD_k^2(p, q) =& \mathbb{E}_p[k(x, x')] \\ &- 2\mathbb{E}_{p,q}[k(x, y)] + \mathbb{E}_q[k(y, y')]. \nonumber
\end{align} 
such that $x, x' \sim p$ and $y, y' \sim q$. 
For finite samples $\mathbf{x} = \{ x_i\}_{i=1}^N$ and $\mathbf{y} = \{ y_i\}_{j=1}^{M}$, the empirical approximation of the MMD can be calculated as
\begin{align}\label{eq:mmd_empirical}
    \MMD_k^2&(p, q) \approx \overline{\MMD}_k^2(\mathbf{x}, \mathbf{y}) = \frac{1}{N^2} \sum_{i=1}^N \sum_{i'=1}^N k(x_i, x_{i'}) \\
    &- \frac{2}{NM} \sum_{i=1}^N \sum_{j=1}^M  k(x_i, y_i) + \frac{1}{M^2} \sum_{j=1}^M \sum_{j'=1}^M k(y_j, y_{j'})  \nonumber.
\end{align}
In this work, we seek to use the MMD metric to define \emph{time-varying} search domains that allow robots to adapt to their search problem. 

\section{Ergodic Search on Flows}
\label{sec:main}

\subsection{Ergodicity Over Flows}

Here, we derive the notion of ergodicity for time-varying domains. We assume a compact static domain $\Omega_0 \subset \espace$ which is enclosed in a larger ambient space $\espace = \R^n$. Furthermore, we consider a probability measure $\mu_0 \in \mathcal{P}(\espace)$ which is supported on $\Omega_0$. In this work, we consider discrete-time varying domains as those that evolve by some induced \emph{discrete flow}, i.e., a function $\omega_{t+1} = F(\omega_{t}) \forall \omega \in \Omega$.
\begin{definition} (Discrete Flow Function)
    A \emph{discrete flow function} is a function $\zeta: \mathbb{Z}^+ \to C(\espace,\espace)$ which satisfies:
    \begin{itemize}
        \item $\zeta_0 = \Id_\espace$ is the identity on $\espace$;
        \item $\zeta_s \circ \zeta_t = \zeta_{s+t}$.
    \end{itemize}
    We say that the discrete flow is \emph{invertible} if $\zeta_t \in \Homeo(\espace)$ for all $t$ (a function $f:\espace \to \espace$ is a \emph{homeomorphism} if $f$ is continuous and has a continuous inverse $f^{-1}$).
\end{definition}

The idea is that the flow induces the changes in both the domain and the underlying information measure. In particular, we set $\Omega_t = \zeta_t(\Omega_0)$, and define the \emph{push-forward measure} $\mu_t \coloneqq (\zeta_t)_* \mu_0$\footnote{In simpler terms, this is just the information measure pushed forward onto the domain at time $t$.} by
\begin{align}
    \mu_t(A) = \mu_0(\zeta_t^{-1}(\mathcal{A}))
\end{align}
for any Borel set $\mathcal{A} \subset \espace$. Note that $\mu_t$ is a probability measure supported on $\Omega_t$. We will now consider the notion of ergodicity with respect to such time-varying measures induced by a flow, beginning with the invertible setting, and then the general setting.\medskip

\noindent \textbf{Invertible Flows.} We can define ergodicity for discrete invertible flows as follows.

\begin{definition} (Time-Varying Ergodicity for Invertible Flows) \label{def:time-varying-erg-invertible}
    Suppose $\zeta$ is a discrete invertible flow. We say that a trajectory $\mathbf{x} = \{x_t\}_{t=0}^{T-1}$ is \emph{ergodic} with respect to the time-varying measure $\mu_t$ if
    \begin{align}
        \lim_{T \to \infty} \frac{1}{T} \sum_{t=0}^{T-1} \phi(\zeta_t^{-1} \circ g \circ x_t) = \int_{\Omega_0} \phi(\omega) d\mu_0(\omega)
    \end{align}
    for all $\phi \in C(\Omega_0)$.
\end{definition}

The invertibility of the flow allows us to map our trajectory $g(x_t)$ back to this original domain. Therefore, this motivates one definition of an MMD metric by mapping the trajectory statistics back to $\Omega_0$, and using the standard MMD metric for $\mu_0$. In particular, we have
\begin{align} \label{eq:init_mmd}
&\overline{\MMD}^2_{k,\zeta}(\rho_\mathbf{x}, \mu) \\
&= \frac{1}{T^2} \sum_{t,t'=0}^{T-1}  k(\zeta_t^{-1} \circ g \circ x_t, \zeta_{t'}^{-1} \circ g \circ x_{t'}) \nonumber \\
        &- \frac{2}{TM} \sum_{t,j}  k(\zeta_t^{-1} \circ g \circ x_t, \omega_j) + \frac{1}{M^2} \sum_{j,j'=1}^M  k(\omega_j, \omega_{j'})  \nonumber,
\end{align}
where all the $\omega_j$'s are sampled from $\mu_0$. \medskip

\begin{figure*}
  \centering  \includegraphics[width=\textwidth]{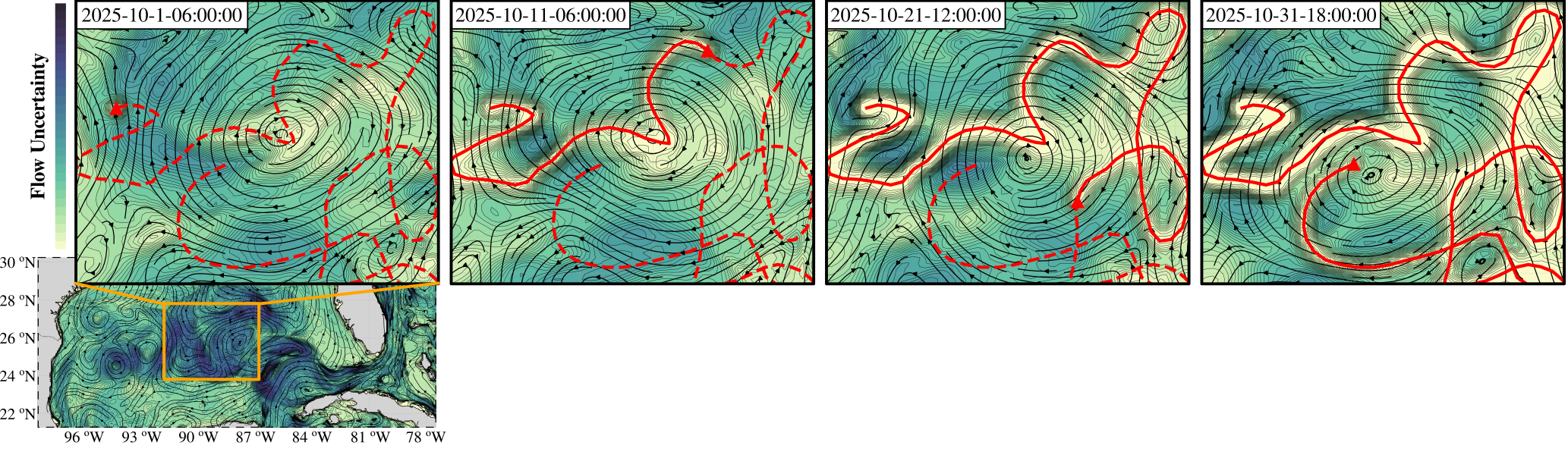}
  \vspace{-65pt}
  \captionsetup{margin={5.45cm,0cm}}
  \captionof{figure}{\textbf{Long-term Modeling of Gulf Stream Ocean Currents.} 
  The utility contour represents areas where flow approximation models disagree.  
  Using the proposed method, an underwater robot visits regions proportional to their  
  utility. The clearing utility contours in sampled regions indicate the robot's 
  success in resolving localized flow model ambiguities.}
  \vspace{-8pt}
  \label{fig:abstract}
\end{figure*}

\noindent
\textbf{General Flows.}
In general, if we consider a non-invertible flow, the heuristic idea of time-varying ergodicity is that the trajectory statistics $\rho_{\bx,T}$ should weakly converge to $\mu_T$ as $T \to \infty$. However, $\rho_{\bx,T}$ contains samples collected at different times, and we need to map these samples forward to time $T$ along the flow in order to compare with the measure $\mu_T$. In particular, we define the \emph{trajectory distribution with respect to the forward flow $\zeta$ at time $T-1$} as
\begin{align} 
    \rho^{\zeta,f}_{\bx,T} \coloneqq \frac{1}{T} \sum_{t=0}^{T-1} \delta[ \omega - \zeta_{T-1-t} \circ g(x_t)].
\end{align}
Now, we can give the general definition of time-varying ergodicity.
\begin{definition} (General Time-Varying Ergodicity)\label{def:time-varying-erg}
    Suppose $\zeta$ is a discrete flow. We say that a trajectory $\mathbf{x} = \{x_t\}_{t=0}^{T-1}$ is \emph{ergodic} with respect to the time-varying measure $\mu_t$ if
    \begin{align}
        \lim_{T \to \infty} \frac{1}{T} \sum_{t=0}^{T-1} \phi(\zeta_{T-t-1} \circ g \circ x_t) - \int_{\Omega_T} \phi(\omega) d\mu_T(\omega) = 0
    \end{align}
    for all compactly supported $\phi \in C_c(\espace)$. 
\end{definition}
Note that with this definition of ergodicity, what matters is what the measure $\mu_T$ looks like as $T \to \infty$. In particular, if the flow is not injective, and identifies two points $x, y \in \Omega_0$; for instance $\zeta_t(x) = \zeta_t(y)$, then this definition of ergodicity cannot distinguish between which of the two points have been visited. 
Note that in the Lipschitz invertible setting, the two definitions are equivalent.
\begin{prop} \label{prop:equivalence_of_erg_def}
    Suppose $\zeta$ is a discrete invertible flow where each $\zeta_t: E \to E$ is a Lipschitz continuous function with Lipschitz constant $\|\zeta_t\|_{\Lip} \leq 1$. Then, Definitions \ref{def:time-varying-erg-invertible} and \ref{def:time-varying-erg} of time-varying ergodicity are equivalent.
\end{prop}

This provides an alternate definition of a time-varying MMD metric which does not involve a backwards map, but rather samples at a later time,
\begin{align} \label{eq:pf_mmd}
&\overline{\MMD}^2_{k,\zeta}(\rho_\mathbf{x}, \mu) \\
&= \frac{1}{T^2} \sum_{t,t'=0}^{T-1}  k(\zeta_{T-t-1} \circ g \circ x_t, \zeta_{T-t'-1} \circ g \circ x_{t'}) \nonumber \\
        &- \frac{2}{TM} \sum_{t,j}  k(\zeta_{T-t-1} \circ g \circ x_t, \omega_j) + \frac{1}{M^2} \sum_{j,j'=1}^M  k(\omega_j, \omega_{j'})  \nonumber,
\end{align}
where now, the $\omega_j$'s are sampled from $\mu_{T-1}$. 
Recall that we do not need to measure ergodicity as $T\to \infty$ and instead only consider finite-time horizon problems. 
Thus, we can use the definitions of MMD to define two formulations of ergodic control for time-varying domains: 1) when the flow of the domain is given over a finite time; and 2) when the transform $\zeta_{t'}$ of the domain at some time $t'$ to its original configuration $\Omega_0$ can be measured (e.g., when we are tracking an object and know where it is at the current time). 

While this formulation enables exploration over moving targets, the discrete samples that move according to a known flow field also can be viewed as a non-stationary distribution where the samples move according to max-likelihood. Hence, the target distribution is inherently non-stationary and assumed to be evolving subject to some flow. 
This allows practitioners to plan for arbitrary transient distributions by constructing a virtual flow field that relocates sample densities over time.

\subsection{Ergodic Trajectory Optimization on Flows}

We now formulate the optimal control problem for generating ergodic trajectories on flows.
    First, let us define the ergodic metric given a general flow $\zeta$ (either forward Def.~\ref{def:time-varying-erg} or backward Def.~\ref{def:time-varying-erg-invertible}) as 
    \begin{align}\label{eq:e_mmd}
        \mathcal{E}^k_\mu(\bx, \zeta) = \overline{\MMD}^2_{k,\zeta}(\rho_\mathbf{x}, \mu)
    \end{align}
    where $\zeta$ encodes some measurable transform or known flow at the current time $t$ that projects the initial information measure $\mu_0$ towards $\mu_t$.
    Given state and control constraints $h_1, h_2$ on the robotic system, e.g., initial condition, control saturation, the optimization problem is defined as 
    \begin{align} \label{prob:emmd}
        \min_{\substack{\mathbf{x}, \mathbf{u}}}\,\, & \mathcal{E}^k_\mu(\bx, \zeta)\\ 
        \text{subject to } & h_1(x_t, u_t) = 0 \,\, \forall t \in [0, T-1] \nonumber \\ 
        & h_2(x_t, u_t) \le 0 \,\, \forall t \in [0,T-1] \nonumber
    \end{align}
    for constraints $h_1(x_t, u_t)$ and $h_2(x_t, u_t)$. 
    Note that in this formulation, we only need information about the flow map (which specifies whether we are looking at future samples from information density $\mu_t$ or pushing back samples over the original space).
    An illustration of the approach is shown in Fig.~\ref{fig:bunny_search_flow}.

\subsection{Computational Complexity} 
    
    Per optimization step, the computational complexity of the flow-adaptive MMD metric is $O(T^2+TM+M^2)$. 
    Because the target distribution's self-similarity term is constant throughout the optimization, we can eliminate this term to reduce the computational scaling to $O(T^2+TM)$.
    This scaling indicates that, for shorter time-horizons, the optimization's performance is primarily sensitive to the spatial sample resolution $M$.
    Conversely, for longer time-horizons, the computational overhead is quadratically dominated by the number of trajectory control knots $T$.
    The summation-based structure of MMD also makes it highly compatible with GPU parallelization, which can reduce the computational scaling to approximately linear with respect to the planning horizon.

    

\section{Results}
\label{sec:results}

This section details simulated and physically implemented results of the proposed flow-adaptive ergodic planner's approach in a variety of scenarios. 
We compare the effectiveness of our method against an existing flow-adaptive information-theoretic trajectory optimizer \cite{subramani2017} and a flow-modified information-maximization approach. 
Additionally, we examine the variation in trajectory behavior between identical flow-adaptive ergodic algorithms when provided with varying levels of actuation. 
Finally, we showcase the success of our method in real-world scenarios including cattle behavior analysis and indoor human search tasks. 
Specifically, we are interested in addressing the following questions:

\begin{itemize}
    \item To what degree does flow-adaptive ergodic planning mitigate the limitations of under-actuated control in high-velocity fields?
    \item What types of flows can the proposed method cover?
    \item Can ergodic planning methods yield robustness to variations in approximated flow fields?
\end{itemize}


\subsection{Comparison to Existing Methods}

\begin{figure}
    \vspace{6pt}
    \centering
    \includegraphics[width=\linewidth]{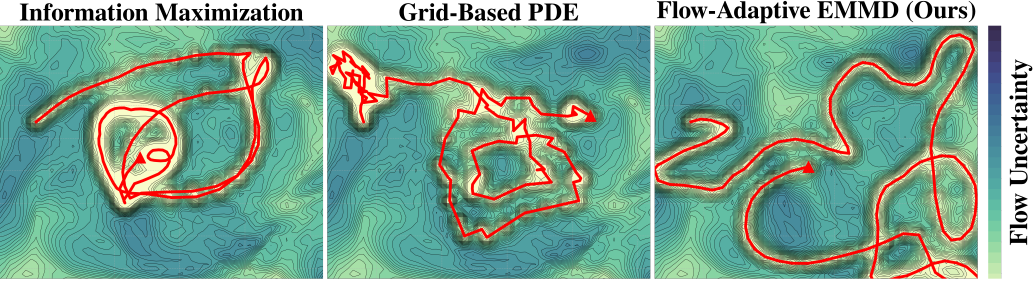}
    \caption{\textbf{Illustrative Trajectories for Eddy Exploration.} A robot is tasked with exploring an eddy subjected to a flow (Fig. \ref{fig:abstract}) to refine a flow model. Each panel displays a robot's trajectory (red line) and the resulting reduction in model uncertainty. 
    Our method leverages the underlying currents to maximize coverage of high-utility regions.}
    \vspace{-10pt}
    \label{fig:compare_2d}
\end{figure}

Here, we demonstrate the proportional coverage effectiveness of our flow-adaptive ergodic planner by comparing our approach to two information-theoretic baselines: 
\begin{itemize}
    \item \textbf{Flow-Modified Information Maximization} \cite{schlotfeldt2019} is a naive implementation of information maximization with integrated flow dynamics that recognizes visitation on moving samples. 
    \item \textbf{Energy-Optimal Planning} \cite{killer2025, subramani2017} is a dynamic programming approach to navigating over uncertain currents towards high-value regions by solving a grid-structured partial differential equation. 
\end{itemize}
Each solver is provided the same optimization parameters, including velocity control bounds of 1.74 knots, and tasked with exploring an eddy within the Gulf of America for one month over 6 hour control intervals (see Fig. \ref{fig:abstract} for eddy location). 
The utility grid shown in Figure \ref{fig:compare_2d} is a normalized distribution of the discrepancy between the true Gulf flow data for the month of October, 2025, against a single-day flow sample collected from August, 2025. 
This data is intended to mimic the discrepancies between existing flow-approximation methods used in the Gulf, and the velocity field used for the robot dynamics is similarly collected for the month of October and changes every 6 hours. 
Given this data, the robot seeks to minimize the overall flow discrepancy by sampling the true flow field every 6 hours during the mission and resolving ambiguities in the flow, as shown by the reduction of contour intensity in regions near the robot. 

As shown in Figure \ref{fig:compare_2d}, the information maximization approach performs moderately well, achieving an overall flow discrepancy reduction of 12.9\%. 
However, the lack of emphasis on wide-spread coverage in the info-max algorithm results in revisitation of previously-seen areas, resulting in sub-optimal performance. 
The energy-optimal solver performed mildly worse than the info-max approach due to its numerical stability's reliance on the satisfaction of the Courant-Friedrichs-Lewy condition, which cannot be guaranteed in all flow-subjected environments \cite{courant1967}. 
Additionally, the feedback policy of this method defines its displacement as a full thrust along the policy gradient, leading to coarse trajectories that may overshoot and miss high-value regions \cite{killer2025}. 
In contrast, the flow-adaptive ergodic planner produces smooth, dynamically-consistent trajectories resulting in a total flow discrepancy reduction of 18.3\%.
The flow-adaptive ergodic objective's simultaneous balance between local information gain and global visitation proportionality allows the robot to anticipate the advection of high-utility features instead of reacting to them, which yields a 42\% improvement over the best-performing baseline.

\begin{table}[htbp]
    \centering
    \label{tab:compare_2d}
    \caption{Eddy Model Refinement Comparison}
    \begin{tabular}{@{}lcc@{}}
        \toprule
        \textbf{Method} & \textbf{Mean Model Error Reduction} \\
        \midrule
        \textbf{Flow-Adaptive EMMD (Ours)} & \textbf{\textbf{18.28} $\pm$ \textbf{3.80} \%} \\
        Flow-Modified Info-Max                           & 12.87 $\pm$ 0.65 \% \\
        Energy-Optimal Planning                      & 11.47 $\pm$ 7.51 \% \\
        \bottomrule
    \end{tabular}
    \vspace{-15pt}
\end{table}

\subsection{Energy-Constrained Exploration}

Next, we evaluate the variation in our methods' coverage effectiveness when provided with varying levels of actuation capacity. 
We construct an environment with 75 randomly-placed samples and generate a vortex flow field that influences a robot and these samples. 
The robot is simulated for 30 different random sample sets and provided with identical optimization parameters except for a varied velocity control bound between 0 m/s and 0.5 m/s. 
When placed in the vortex, where the maximum flow speed is 3.46 m/s and the average flow speed is 2.56 m/s, the robot is tasked with maximizing the number of points visited within 10 seconds.

\begin{figure}[h]
    \centering
    \includegraphics[width=\linewidth]{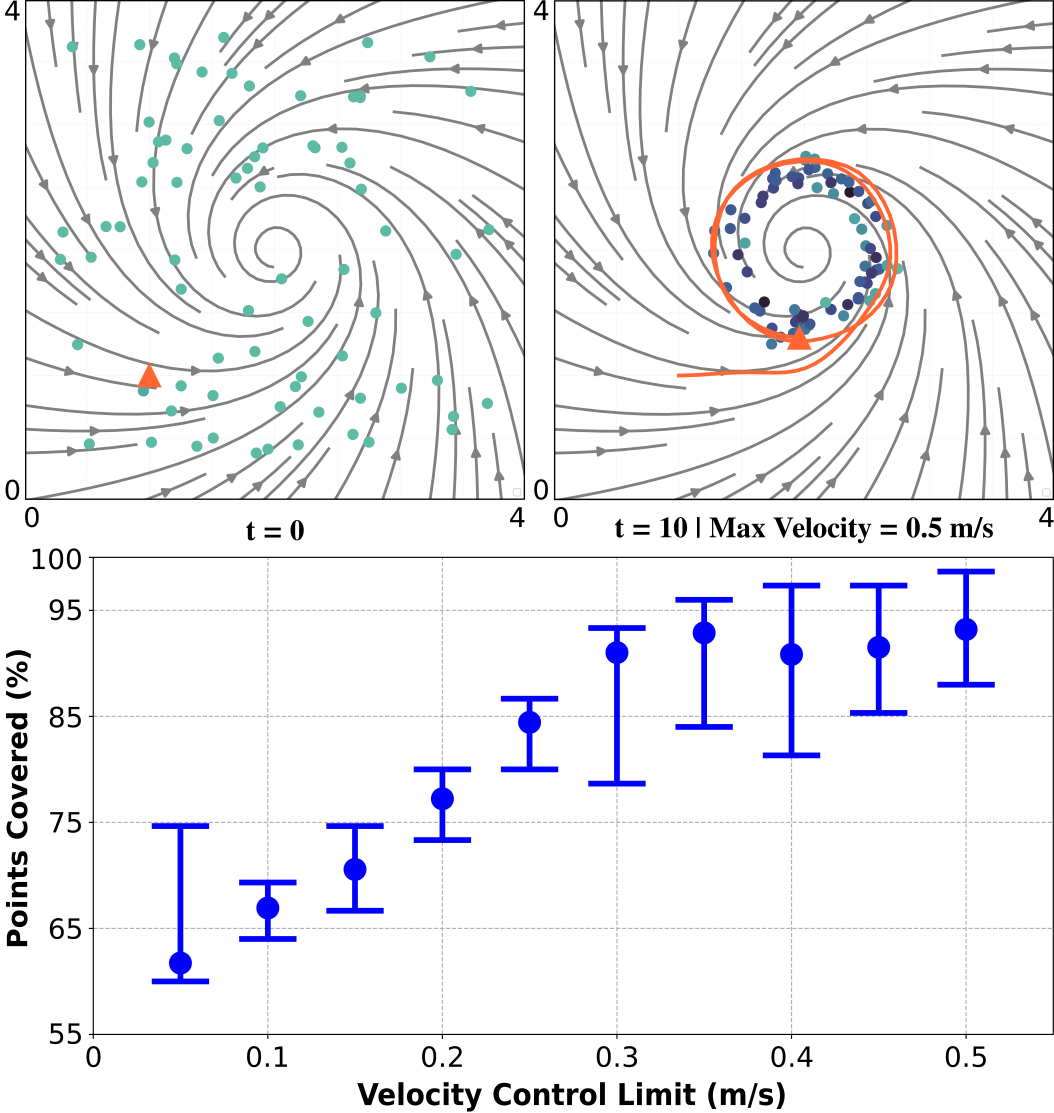}
    \caption{\textbf{Ablation Study on Coverage Effectiveness Under Actuation Constraints.} (Top) Snapshots indicate an initialization of randomly-placed samples within a vortex field with a peak speed of 3.46 m/s and a mean speed of 2.56 m/s. Dark blue color denotes visitation by the trajectory. (Bottom) Performance as a function of velocity control limits. Even at lower control limits, the proposed method is able to leverage the flow field to assist with visiting particles that evolve on the flow.}
    \vspace{-15pt}
    \label{fig:ablation_power}
\end{figure}

When provided with the maximum velocity control bound of 0.5 m/s, approximately 19\% of the average flow speed, the flow-adaptive ergodic planner enables the robot to visit up to 98.67\% of spatial samples within 10 seconds (see Fig. \ref{fig:ablation_power}).
As shown in in Figure \ref{fig:ablation_power}, the coverage quality of the robot's trajectories steadily increases as the robot's velocity bound increases and the robot gains the ability to maneuver through the flow. 
While the flow-adaptive ergodic planner is inherently limited by a given robot's ability to traverse a given flow field, even in under-actuated systems, our method plans trajectories that leverage the flow field and the robot's limited actuation to work with the flow and maximize visitation. 


\subsection{Experimental Validation}

\begin{figure}[t]
    \centering
    \includegraphics[width=\linewidth]{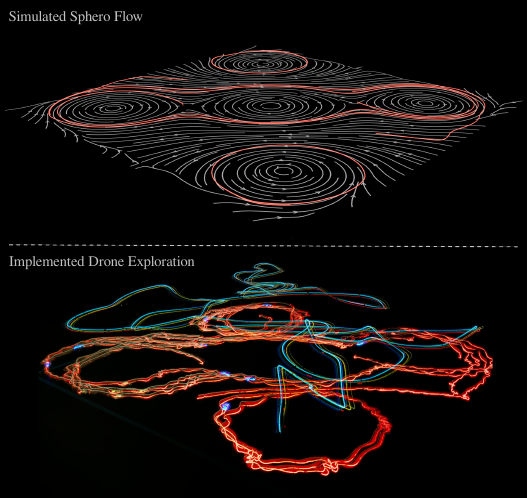}
    \vspace{-12pt}
    \caption{\textbf{Experimental Deployment in a Simulated Flow-Field.} (Top) Simulated flow-following trajectories for a fleet of 8 Spheros. (Bottom) Corresponding hardware implementation using a Crazyflie 2.1 drone (blue trail) and a Sphero fleet (red trails) over 40 seconds. The proposed method effectively manages the drone's flight dynamics to maintain ergodic coverage over target distributions subjected to time-varying flow patterns and sample sparsity.}
    \vspace{-5pt}
    \label{fig:sphero_example}
\end{figure}

\noindent
\textbf{Real-World Exploration Using Crazyflie Drone.} 
First, we demonstrate our system's compatibility to robot dynamics and sparse information distributions by performing a monitoring task on a fleet of 8 Spheros using a Crazyflie 2.1 drone. 
The Spheros are distributed within a 3m x 3m space and instructed to follow trajectories that imitate each Sphero's movement through a flow field for 40 seconds (shown in Fig. \ref{fig:sphero_example}). 
A Crazyflie drone is then instructed to traverse the space, dynamically independent of the flow, to observe each of the Spheros and revisit Spheros once each has been seen. 
Our method successfully plans a coverage trajectory that visits each Sphero within 30 seconds while respecting drone dynamics. 
The MMD-based ergodic metric's use of discrete spatial samples to represent a continuous distribution allows for trajectory optimization at varying levels of sample density, even in systems of extreme sparsity (e.g., 8 samples). 

\begin{figure}[t]
    \centering
    \includegraphics[width=\linewidth]{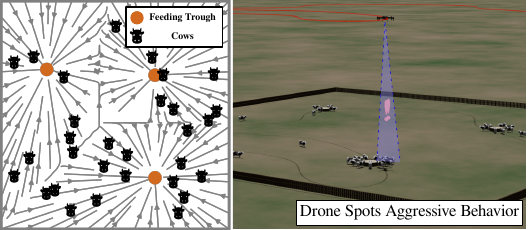}
    \vspace{-12pt}
    \caption{\textbf{Validation of Ergodic Coverage In a Behavioral Flow Environment.} 
    (Left) Thirty cows are shown in a fenced-in region at t = 0, where gray vectors indicate the directed vector field towards the nearest feeding trough to each cow. (Right) A drone executes an ergodic trajectory designed to prioritize regions of high cattle density to locate potential conflicts between cattle. Our method successfully positions the drone over problematic cows (marked with exclamation points) during feeding time. We show that the proposed method generalizes beyond fluid currents to any motion-predictable system.}
    \vspace{-15pt}
    \label{fig:cow_feeding}
\end{figure}

\noindent
\textbf{Simulated Use-Case of Cattle Behavior Inspection.}
When obtaining new cattle, farmers often need to observe the interactions between newly-acquired cattle and the existing herd. 
A common point of contention for more aggressive cattle is during feeding time, when the herd movement is approximated according to a given cow's proximity to each feeding spot.

We evaluate our method's effectiveness at anomaly detection by monitoring aggressive interactions during cattle feeding. 
As shown in Fig. \ref{fig:cow_feeding}, we conduct 30 simulations involving 30 cows randomly distributed across a 30m $\times$ 30m ranch with three feeding troughs (flow attractors). 
In each trial, two cows are selected as aggressive and a drone with a 8m sensing radius is tasked with identifying conflicts involving aggressive cows.

The reliance of the energy-based planner on the Courant-Friedrichs-Lewy (CLF) condition, which requires that the numerical wave speed of the PDE is at least as fast as the physical wave speed, results in unstable numerical solutions during the path-planning process due to the high-velocity gradients required to model the herd's behavioral flow. 
In contrast, our flow-adaptive ergodic planner successfully guides the drone to aggressive interactions with a 50.8\% success rate in an open-loop setting (see Tab. \ref{tab:cattle-results}). 
While information maximization tends to see coverage improvements in search problems with sparse data (e.g., 30 cows), our method balances prioritization of high-utility visitation and wide-spread coverage, resulting in a 42.8\% improvement in performance.
The success of this use-case demonstrates that the flow-adaptive ergodic planner enables exploration over predictable behavioral flows in addition to fluid flow.

\begin{table}[htbp]
    \centering
    \caption{Open-Loop Cattle Skirmish Detection Comparison}
    \label{tab:cattle-results}
    \begin{tabular}{@{}lcc@{}}
        \toprule
        \textbf{Method} & \textbf{Cows Seen} & \textbf{Success (\%)} \\ 
        \midrule
        Total Cattle Fights               & 59 &  \\ 
        \hdashline \noalign{\smallskip}
        \textbf{Flow-Adaptive EMMD (Ours)} & \textbf{30} & \textbf{50.847} \\
        Flow-Modified Info-Max \cite{schlotfeldt2019}                          & 21          & 35.593          \\
        Energy-Optimal Planning \cite{killer2025}                     & {\color{red} \faTimes}    & {\color{red} \faTimes}       \\
        \bottomrule
    \end{tabular}
\end{table}

\begin{figure}
    \centering
    \includegraphics[width=\linewidth]{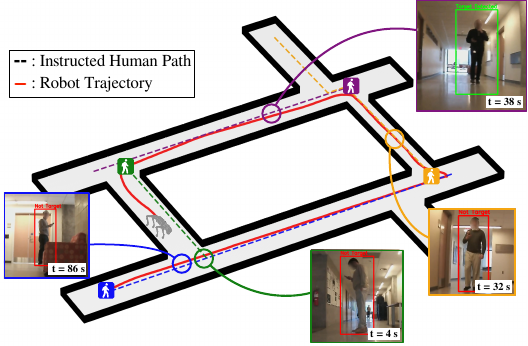}
    \vspace{-12pt}
    \caption{\textbf{Implementation of Flow-Adaptive Ergodic Coverage to Indoor Search Tasks.} A Unitree Go2 quadruped robot (red trajectory) is tasked with locating a person of interest (wearing all black) among a group of four people. The motion of each person is modeled as a constrained social flow. The robot's camera view confirms the detection of the target during exploration in loose approximations of social flows.
    }
    \vspace{-15pt}
    \label{fig:go2}
\end{figure}

\noindent
\textbf{Indoor Human Search on Quadruped Robot.}
In architecturally-constrained environments, such as hallways, human motion is similarly approximable as a flow. 
To test our flow-adaptive ergodic planner on motion patterns with less adherence to the instructed flow, we deploy a Unitree Go2 quadruped robot with an approximated flow for 4 people walking through a floor of an office building. 
Each person is instructed to follow a movement pattern (shown in Fig. \ref{fig:go2}) which is provided to the Go2's trajectory planner to perform uniform search over the 4 people to locate the person wearing an all-black outfit. 
To test the robustness of our solver, each person was instructed to introduce stochastic deviations, such as lateral drifting or mid-path stops, rather than following the path perfectly. 
For additional implementation details, please see Appx. A. 

As shown in Figure \ref{fig:go2}, the Go2 successfully maneuvered the indoor non-convex domain to observe each person on the floor and locate the target, with the robot successfully locating the target during the 2-minute run-time. 
The kernel-based spatial integration inherent to MMD provides a form of stochastic smoothing that yields resilience to minor flow model deviations, such as those made by humans following a path.
With this validation, we show that our flow-adaptive planner successfully plans trajectories that adhere to high-dimensional robot dynamics while maintaining sufficient robustness to locate a human target who performs minor deviations from their instructed movement pattern. 
The success of this experiment validates that architecturally-imposed motion constraints (e.g., hallways) can create pseudo-deterministic flows that are sufficiently structured for ergodic coverage planning. 

\noindent
\textbf{Simulated Whale Tracking on Real-World Gulf Data.} 
Finally, we perform a simulated validation of our method in the Gulf stream where an underwater robot with 1.16 knots of actuation ability is assigned to inspect 6 whales traveling with the ocean's current. 
As shown in Fig. \ref{fig:whales}, our approach successfully observes each whale over the course of a 1-month migration while leveraging the ambient flow to ensure visitation before the pod separates. This validates our method's consideration of the appropriate time to visit each sample in the search space by sometimes delaying short-term visitation to maximize long-term proportionality. 


\section{Limitations}
\label{sec:limitations}

The primary limitation of the proposed approach is its dependence on the fidelity of the environmental flow model in open-loop. 
While this reliance is negligible in domains that evolve through known flows, large stochastic environments, such as oceanic currents or unconstrained crowds, introduce inherent model uncertainties. 
This model uncertainty presents two distinct errors in our trajectory planning: coverage degradation and state-estimation error, which lead to short-term performance sensitivities. 

Theoretically, ergodic search methods are robust to modeling inaccuracies over infinite time-horizons, but would require excessive computation. A recursive online adaptation, like MPC with flow model correction, could resolve this issue, but its impact on coverage guarantees and how one would measure success in a MPC-like setting require significantly more analysis beyond the scope of this work.



\section{Conclusion} 
\label{sec:conclusion}

In this paper, we present a novel formulation of the coverage-based exploration problem as an ergodic trajectory optimization solved over time-varying flows. 
We derive the optimization objective for ergodic exploration over general flows using the maximum mean discrepancy (MMD) metric and show that coverage guarantees over the reachable search space are maintained despite the influence of flow dynamics in open-loop planning. 
Our results demonstrate that our approach outperforms traditional information-theoretic baselines by at least 42\% while maintaining high-fidelity coverage in under-actuated problem settings without the need for replanning. 

The effectiveness of our proposed approach is demonstrated in highly restrictive oceanographic applications that are unable to perform onboard computation. Additionally, our approach allows autonomous platforms to leverage the ambient currents to improve mobility and intercept high-value features that are advected through the search space. 
Last, we show that the proposed method generalizes to behavioral flows, such as general human traffic patterns and herd movement, by treating architectural constraints or universal attractors as pseudo-deterministic flow fields. 






\bibliographystyle{plainnat}
\bibliography{references}

\end{document}